\documentclass[letterpaper, 10 pt, conference]{ieeeconf}  

\IEEEoverridecommandlockouts                              
\usepackage{graphics} 
\usepackage{epsfig} 
\usepackage{mathptmx} 
\usepackage{times} 
\usepackage{amsmath} 
\usepackage{amssymb}  
\usepackage{multirow}
\usepackage{color}
\usepackage{balance}
\usepackage{amsmath,bm}
\usepackage{caption}
\allowdisplaybreaks

\title{\LARGE \bf
Legged Robot State Estimation in Slippery Environments Using Invariant Extended Kalman Filter with Velocity Update
}

\author{Sangli Teng, Mark Wilfried Mueller, and Koushil Sreenath
\thanks{This work was supported in part by National Science Foundation Grants IIS-1834557, CMMI-1944722 and Berkeley Deep Drive.}
\thanks{Sangli Teng is with the Robotics Institute at University of Michigan, MI 48103, USA, {\tt\small sanglit@umich.edu}}%
\thanks{Mark Wilfried Mueller, and Koushil Sreenath are with the Department of Mechanical Engineering at University of California, Berkeley, CA 94720, USA, {\tt\small \{mwm, koushils\}@berkeley.edu}}%
}

\begin{document}

\setlength{\abovedisplayskip}{0pt}
\setlength{\belowdisplayskip}{0pt}%
\setlength{\textfloatsep}{0pt}	
\setlength{\belowdisplayshortskip}{0pt}

\maketitle 
\thispagestyle{empty}
\pagestyle{empty}

\begin{abstract}
This paper proposes a state estimator for legged robots operating in slippery environments. An Invariant Extended Kalman Filter (InEKF) is implemented to fuse inertial and velocity measurements from a tracking camera and leg kinematic constraints. {\color{black}The misalignment between the camera and the robot-frame is also modeled thus enabling auto-calibration of camera pose.} The leg kinematics based velocity measurement is formulated as a right-invariant observation. Nonlinear observability analysis shows that other than the rotation around the gravity vector and the absolute position, all states are observable except for some singular cases. Discrete observability analysis demonstrates that our filter is consistent with the underlying nonlinear system. An online noise parameter tuning method is developed to adapt to the highly time-varying camera measurement noise. The proposed method is experimentally validated on a Cassie bipedal robot walking over slippery terrain. A video for the experiment can be found at https://youtu.be/VIqJL0cUr7s.

\end{abstract}

\section{Introduction}


For legged robots to navigate complex environments with slippery and unstable terrain illuminated with poor light, state estimation becomes important. To enable navigation in complex environments, state estimators for legged robots that fuse measurements from a wide range of sensors, such as inertial, contact and visual information are needed.

Typically, filters such as Unscented Kalman Filter (UKF), Extended Kalman Filter (EKF) and Invariant Extended Kalman Filter (InEKF) are used to fuse inertial measurements with leg kinematics for legged robot state estimation \cite{ETHslipUKF,ETHSlipPrior,humanEKFusc,hartley2020contact}. These methods exhibit good performance when the contact point is static, as assumed in their system models. However, the drawback of these results is the degraded estimates when the contact point slips due to either slippery ground or unstable terrain. Potential slippage in these early research is either treated as noise, see  \cite{ETHSlipPrior, hartley2020contact, HartleyVIFactor}, or the measurement with slippage is considered as outliers that are rejected in the update, see \cite{ETHslipUKF}. If the slippage is consistent or has a large magnitude, these methods can fail. 
\begin{figure}
  \centering
  \includegraphics[width=0.8\columnwidth]{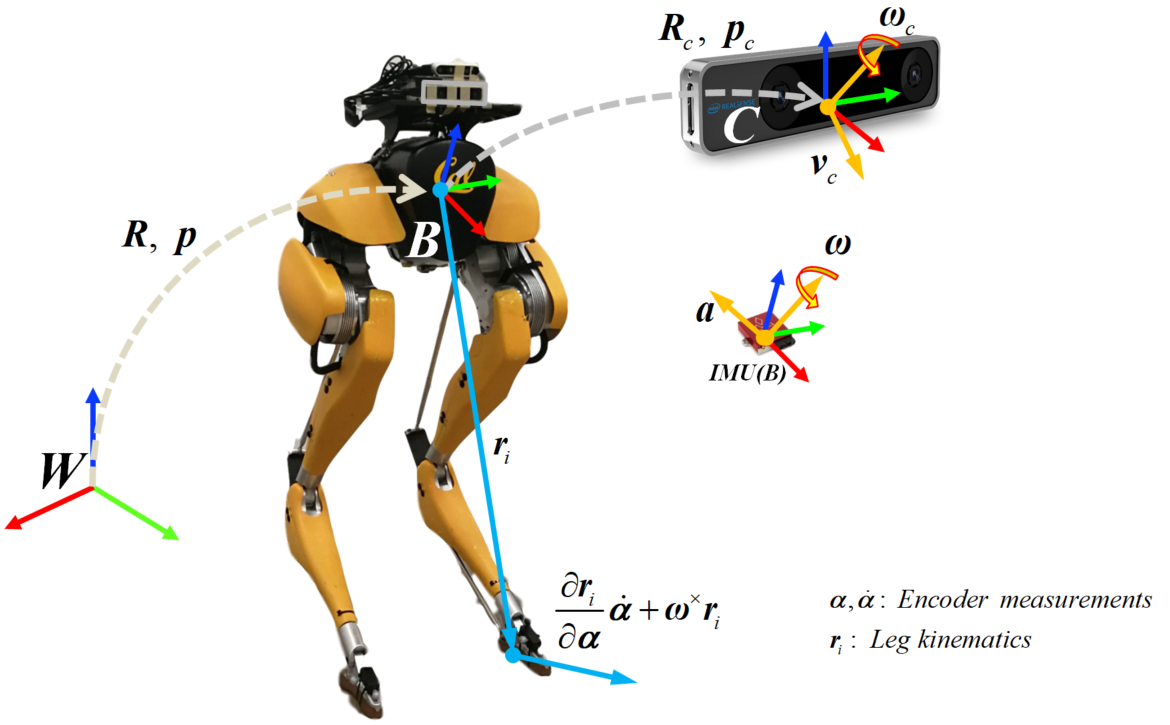}
  \caption{The Cassie bipedal robot used for the experiment. Orientation $\bm{R}$ and position $\bm{p}$ of robot pelvis (IMU) are represented with respect to the world frame ($\bm{W}$). The tracking camera pose, i.e. $\bm{R}_c$ and $\bm{p}_c$ are expressed with respect to the robot-frame ($\bm{B}$). The tracking camera mounted on the robot's pelvis can provide velocity ${\bm{v}}_c$ and rotational rate ${\bm{\omega}}_c$ measurements in camera pose frame ($\bm{C}$).} 
  \label{robot_camera_system}
\end{figure}
Visual-inertial based methods have also been explored for state estimation. Research in \cite{ETHirosOpticalFlow} uses optical flow measurement for UKF innovation, where the FAST corner detector, \cite{FAST}, serves as the underlying algorithm to extract features. Research in \cite{ETHijrrVIO} directly integrates the landmarks to the state space. However, fast lighting changes or loss of texture can greatly influence the quality of landmark extraction, resulting in bad state estimation. 

Research in \cite{hartleyVICfactor} and \cite{oxfordAnymalVICfactor} use a factor graph based method that integrates inertial, leg kinematic and visual information for long range navigation in legged robots. Both researches reveal that inertial-kinematic-visual methods outperform the methods with less information fused. However, since the factor graph based methods use measurements along the whole trajectories for smoothing, the update-rate is too low for real-time control purposes. 

{\color{black}To address slippery environments, \cite{slipfloor} and \cite{slipAcc} add force sensors or accelerometers, respectively, on the stance leg to detect slip events. However, these sensors are vulnerable to ground impact. A probabilistic method is used in \cite{ETHSlip} for slippage and contact detection. Contact state obtained by \cite{ETHSlip} is fused to the filter in \cite{ETHslipUKF} for outlier rejection. A Constraint Kalman Filter is adopted in \cite{conEKF} to include the linear complementary condition involved in contact modes. The external forces, friction coefficient and contact states are determined by nonlinear optimization techniques. However, information of the terrain is required to parameterize the external force, e.g. the slope of the ground, which may restrict the use of the method.} 

In this paper, we develop a filter-based approach that addresses the state estimation for legged robot walking over slippery terrain in a computationally efficient manner. The key idea is to fuse the vision-based velocity measurement with inertial and leg kinematics information. In particular, we use the Intel Realsense T265 tracking camera that can provide velocity and rotational rate measurements in the camera frame. The contributions of this paper are:
{\color{black}
\begin{itemize}
    \item Derivation of an InEKF for the inertial-legged-robot-camera system. Measurements model for legged kinematics and tracking camera are presented. 
    \item Auto calibration of camera pose, i.e, estimating the offset between the camera-frame and the robot-frame as part of the state estimation. 
    \item Online noise parameter tuning to adapt to the time-varying manner of camera noise due to ground impact.
    \item Nonlinear and discrete observability analysis. The analysis demonstrates the consistency of our observer.
    \item Experimental validation using a 3D bipedal robot over slippery terrain. Experiment suggests that our method is robust to significant amount of slippage. 
\end{itemize}

The proposed method builds off the inertial-kinematics based work in \cite{ETHslipUKF} and \cite{hartley2020contact}. However, in comparison to these priori work, our method is more robust to consistent slippage, while \cite{hartley2020contact} could diverge and \cite{ETHslipUKF} fails to reject all the slippage outliers thus leading to inaccurate estimation. }

The remainder of this paper is organized as follows. Section II presents system and measurement models. Section III presents the setup of our filter. Observability analysis is presented in Section IV. Section V presents experimental result and conclusions are presented in Section VI.

\section{System and Measurements Model}
For a legged robot, we wish to estimate the orientation $\bm{R}\in SO(3)$, velocity $\bm{v}\in\mathbb{R}^{3}$ and position $\bm{p}\in\mathbb{R}^{3}$ of the robot pelvis in the world frame via the measurements through the inertial measurement unit (IMU), leg kinematics and a tracking camera. The IMU is installed in the robot pelvis and we assume the IMU frame is aligned with the robot pelvis frame. In this section we introduce the model of the robot-camera system and the derive the measurement model. The robot system and the variables are illustrated in Fig. \ref{robot_camera_system}.
\subsection {System Model}
The IMU measures the acceleration $\bm a$ and angular velocity $\bm \omega$ in robot (IMU) frame. The measurements  $\tilde{\bm a}$ and $\tilde{\bm \omega}$ are corrupted by white Gaussian noise $\bm{w}_{a}$, $\bm{w}_{\omega }$ and bias $\bm{b}_{a}$, $\bm{b}_{\omega }\in\mathbb{R}^{3}$:
\begin{equation}
      \bm{\tilde{\omega}}={\bm{\omega }}+\bm{b}_{\omega }+\bm{w}_{\omega },\ \ \ \  {{{\bm{\tilde{a}}}}}={{\bm{a}}}+\bm{b}_{a}+\bm{w}_{a}.
\end{equation}
The IMU bias are modeled as random walk, i.e, their derivatives are white Gaussian noise $\bm{w}_{{b}{a}}$ and $\bm{w}_{{b}{\omega}}$, i.e.
\begin{equation}
\label{bias}
    \dot{\bm{b}}_{\omega } = \bm{w}_{b{\omega}},\ \ \ \ \dot{\bm{b}}_{a} = \bm{w}_{{b}{a}}.
\end{equation}
Therefore, the dynamics of robot IMU are governed by:
\begin{equation}
    \dot{\bm{R}} = \bm{R}\left( \bm{\tilde{\omega}} - \bm{b}_{\omega } - \bm{w}_{\omega } \right)^{\times}, 
\end{equation}
\begin{equation}
   \dot{\bm{p}} = \bm{v},\ \ \ \dot{\bm{v}} = \bm{R}\left( \bm{\tilde{a}} - \bm{b}_{a} - \bm{w}_{a} \right)+\bm{g}.
\end{equation}
where $\bm{g}$ is the acceleration due to gravity and {\color{black}$({\cdot})^{\times}$ denote a $3\times 3$ skew symmetric matrix, s.t $x\times y = x ^{\times}y, \forall x,y\in \mathbb{R}^{3}$.}

The position and orientation of the tracking camera with respect to the robot-frame could be represented by $\bm p_c\in\mathbb{R}^{3}$ and $\bm R_c\in SO(3)$ respectively. We model the camera pose as constants corrupted by white Gaussian noise $\bm{w}_{pc}$ and $\bm{w}_{Rc}$:
\begin{equation}
\label{camera_pose}
    \dot{\bm{R}}_c=\bm{R}_c\bm{w}_{Rc}^{\times},\ \ \ \ \dot{\bm{p}}_c=\bm{w}_{pc}.
\end{equation}
\subsection {Leg Kinematics Measurements}
The legged robot is equipped with encoders in leg joints that provide us with the corresponding angular position ${\bm \alpha}$ and its derivative $\dot{{\bm \alpha}}:=\frac{\partial {\bm \alpha}}{\partial t}$. The corresponding measurements $\tilde{\bm \alpha}$ and $\dot{\tilde{\bm \alpha}}$ are corrupted by white Gaussian noise. Using leg kinematics, we can compute the location of the $i_{\text{th}}$ contact point $\bm{r}_i(\bm{\alpha})$ in root frame. Let the position of the $i_{\text{th}}$ contact point in world frame be  $\bm{d}_i$, expressed as:
\begin{equation}
    \bm{d}_i=\bm{p}+\bm{R}\bm{r}_i(\bm{\alpha}).
\end{equation}
Taking derivative on both sides, we obtain the velocity of the $i_{\text{th}}$ contact point:
\begin{equation}
    \label{fk_world}
    \dot{\bm{d}}_i=\bm{v}+\bm{R}\bm{J}_i\dot{\bm{\alpha}}+\bm{R}\bm{\omega}^{\times}\bm{r}_i(\bm{\alpha}),
\end{equation}
where $\bm{J}_i =  \frac{\partial \bm{r}_i}{\partial \bm{\alpha}}$. Representing (\ref{fk_world}) in robot-frame, we have:
\begin{equation}
    \bm{R}^{\intercal}\dot{\bm{d}}_i=\bm{R}^{\intercal}\bm{v}+\bm{J}_i\dot{\bm{\alpha}}+\bm{\omega}^{\times}\bm{r}_i(\bm{\alpha}).
\end{equation}

If we assume the $i_{\text{th}}$ contact point remains static to the world frame (no slip), we obtain the measurement model:
\begin{equation}
\label{fk_model}
\begin{aligned}
 \bm{R}^{\intercal}\dot{\bm{d}}_i&=0,\ \ \ \ 
      -\bm{J}_i\dot{\tilde{\bm{\alpha}}}-\tilde{\bm{\omega}}^{\times}\bm{r}_i(\tilde{\bm{\alpha}}) = \bm{R}^{\intercal}\bm{v}+\bm{n}_{f},
\end{aligned}
\end{equation}
where $\bm{n}_{f}$ is white Gaussian noise. For simplification, the single parameter $\bm{n}_{f}$ incorporates the uncertainty in encoder measurements, kinematic model, the effect of slip and gyroscope bias. Similar measurement model, with the gyroscope bias incorporated, is used in \cite{ETHslipUKF} for UKF. {\color{black}However, we will later show that our model could formulate a right invariant observation \cite{barrau2016invariant}.}

\subsection{Tracking Camera Measurements}
The tracking camera mounted on the top of robot could provide the measurements of velocity $\bm{v}_c$ and rotational rate $\bm{\omega}_c$ in the camera-frame. The velocity of the robot represented in the camera-frame is:
\begin{equation}
    \bm{v}_{c}=\bm{R}^{\intercal}_{c}\bm{R}^{\intercal}\bm{v}+\bm{\omega}_{c}^{\times}\bm{R}^{\intercal}_{c}\bm{p}_c.
\end{equation}

{\color{black}We assume the biases of angular velocities and rotational rate measurement have been eliminated by the tracking camera's internal visual inertial odometry algorithm. Therefore, the tracking camera measurements are modeled as solely corrupted by white Gaussian noise $\bm{n}_{v c}$ and $\bm{n}_{\omega c}$}:
\begin{equation}
    \tilde{\bm{v}}_c=\bm{v}_c+\bm{n}_{vc},\ \ \ \ \tilde{\bm{\omega}}_c=\bm{\omega}_c+\bm{n}_{\omega c}.
\end{equation}
The measurements finally becomes:
\begin{equation}
    \label{vel_model}
    \tilde{\bm{v}}_{c}=\bm{R}^{\intercal}_{c}\bm{R}^{\intercal}\bm{v}+(\tilde{\bm{\omega}}_c-\bm{n}_{\omega c})^{\times}\bm{R}^{\intercal}_{c}\bm{p}_c+\bm{n}_{vc}.\\
\end{equation}
\subsection {Problem Formulation}
Based on the system model (\ref{bias}-\ref{camera_pose}) and measurement model (\ref{fk_model}, \ref{vel_model}), we now formulate the filtering problem. For a legged robot, with one tracking camera mounted on the pelvis and leg kinematics constraints, we define the state variables:
\begin{equation}
\label{filter_states}
    \bm{x}:=\left( \bm{R},\ \bm{v},\ \bm{p},\ \bm{b}_{\omega},\ \bm{b}_{a},\ \bm{R}_c,\ \bm{p}_c \right).
\end{equation}
Our goal is to use the measurements from the IMU, leg kinematics and tracking camera to estimate $\bm{x}$. 
{\color{black}Note, by including $\bm{R}_c$ and $\bm{p}_c$ into the state space, we will avoid the problem requiring an accurate calibration of camera pose and thus increase the robustness. }




\section{Invariant Extended Kalman Filter Setup}
The InEKF exploits the symmetry of a system represented by a matrix Lie group. We first provide the prerequisite knowledge of InEKF and then apply it to our system. {\color{black}The InEKF theory and notation are presented in \cite{barrau2016invariant, hartley2020contact}, while the InEKF augmentation of IMU bias are in \cite{barrau2015non, hartley2020contact}, and the system discretization is in \cite{hartley2020contact}.}
\subsection{Math Prerequisite} 
Consider a $n\times n$ matrix Lie group $\mathcal{G}$ and its associated Lie algebra $\mathfrak{g}$, representing the tangent space of $\mathcal{G}$ at the identity. We define the linear map:
\begin{equation}
    (\cdot)^{\land}:\mathbb{R}^{n} \rightarrow \mathfrak{g},
\end{equation}
such that $\forall\bm{\xi} \in \mathbb{R}^{n}$, we can associate it with the elements in the matrix Lie group through the exponential map:
\begin{equation}
    \exp(\cdot):\mathbb{R}^{n} \rightarrow \mathcal{G},\ \ \exp(\bm{\xi})=\operatorname{exp_m}(\bm{\xi}^{\land}),
\end{equation}
where $\operatorname{exp_m}(\cdot)$ is the matrix exponential.

For conversion from local to global coordinate, we define the adjoint map $\mathrm{Ad}_{\bm{\chi}}: \mathfrak{g}\rightarrow \mathfrak{g}$:
\begin{equation}
\label{adjointmap}
    \mathrm{Ad}_{\bm{\chi}}(\bm{\xi}^{\land})= \rm{\bm{\chi}}\bm{\xi}^{\land}\rm{\bm{\chi}}^{-1},\ \ \forall \bm{\chi}\in \mathcal{G}.
\end{equation}

The introduction of the invariant error is the core of InEKF theory. Consider a system defined on matrix Lie group $\mathcal{G}$ and its associate Lie algebra $\mathfrak{g}$ with input $u_t$:
\begin{equation}
\label{system_on_lie_g}
    \frac{\mathrm{d}}{\mathrm{d} t} \bm{\chi}_t=f_{u_t}\left(\bm{\chi}_t\right), \ \ \ \ \ \ \bm{\chi}_t \in \mathcal{G}\ \ t \geq 0. 
\end{equation}

Consider trajectories ${\bm{\chi}}_t$ and $\bar{\bm{\chi}}_t$, the right invariant error between the two trajectories are defined as:
\begin{equation}
\label{inv_error}
\begin{aligned}
\begin{array}{l}
{\bm{\eta}_t=\bar{\bm{\chi}}_t \bm{\chi}_t^{-1}}. \\ 
\end{array} 
\end{aligned}
\end{equation}
Theorem 1 of \cite{barrau2016invariant} suggests that if the system in (\ref{system_on_lie_g}) is group affine, i.e:
\begin{equation}
\label{fut}
f_{u_t}\left(\bm{\chi}_{1} \bm{\chi}_{2}\right)=f_{u_t}\left(\bm{\chi}_{1}\right) \bm{\chi}_{2}+\bm{\chi}_{1} 
f_{u_t}\left(\bm{\chi}_{2}\right)-\bm{\chi}_{1} f_{u_t}\left(\bm{I}_{d}\right) \bm{\chi}_{2},\\
\end{equation}
then the error dynamics are independent of state and satisfy:
\begin{equation}
\label{compute_gut}
\begin{aligned}
     {\frac{\mathrm{d}}{\mathrm{d} t} \bm{\eta}_{t}=}{g_{u_{t}}\left(\bm{\eta}_t\right):=f_{u_{t}}\left(\bm{\eta}_t\right)-\bm{\eta}_t f_{u_{t}}\left(\bm{I}_{d}\right)}. \\ 
\end{aligned}
\end{equation}
Linearizing (\ref{compute_gut}) using first order approximation of the exponential map, we have:
\begin{equation}
\label{linearize_eta}
    \bm{\eta}_t=\operatorname{exp}(\bm \xi_t)\approx\bm{I}+\bm \xi^{\land}_t,
\end{equation}
\begin{equation}
\label{linearize_g}
   g_{u_{t}}(\exp (\bm{\xi}_t)) = \left(\bm{A}_{t} \bm{\xi}_t\right)^{\land}+ o\left(||\bm{\xi}_t||\right) \approx \left(\bm{A}_{t} \bm{\xi}_t\right)^{\land},
\end{equation}
where the the Jacobian matrix $\bm{A}_t$ is independent of $\bm{\xi}_t$.
Combining (\ref{compute_gut}-\ref{linearize_g}), we obtain a linear differential equation:
\begin{equation}
\label{ode}
\frac{\mathrm{d}}{\mathrm{d} t} \bm{\xi}_{t}=\bm{A}_{t} \bm{\xi}_{t}.
\end{equation}



The log-linear property of error propagation \cite{barrau2016invariant} suggests that if the initial error satisfies $ \bm{\eta}_{0}=\exp \left(\bm{\xi}_{0}\right)$, the linearized error dynamics (\ref{ode}) can fully represent the nonlinear error defined in (\ref{inv_error}), as:
\begin{equation}
    \bm{\eta}_{t}=\exp \left(\bm{\xi}_{t}\right),\ \ \ t \geq 0.
\end{equation}

In general, for systems satisfying the group affine property, by the introduction of the invariant error and the log-linear property of error propagation, we can use a linear differential equation (\ref{ode}) to describe the error between two states thus avoiding problems involved in nonlinear observer design. 
\subsection {Estimation Error Dynamics}
Based on the system model (\ref{bias}-\ref{camera_pose}), we give the definition of the error between the estimated states $\hat{\bm{x}}$ and the real states ${\bm{x}}$, where $\hat{({\cdot})}$ denotes the estimated states.

The states of IMU $\left( \bm{R},\ \bm{v},\ \bm{p} \right)$ are embedded in the double direct spatial isometry $SE_{2}(3)$ \cite{barrau2016invariant}: 
\begin{equation}
    \bm{\chi} :=\left[ \begin{matrix}
   \begin{matrix}
   \bm{\ \ R} & \bm{v} & \bm{p}  \\
\end{matrix}  \\
   \begin{matrix}
   \bm{0}_{1\times 3} & 1 & 0  \\
   \bm{0}_{1\times 3} & 0 & 1  \\
\end{matrix}  \\
\end{matrix} \right].
\end{equation}
This matrix is the extension of special Euclidean group $SE(3)$ with additional spatial vectors. Without bias, the dynamics of IMU states are group affine. Thus, we define the right-invariant error of the IMU states and linearize it: 
\begin{equation}
\label{right_invariant_error}
\small{
\begin{aligned}
     \bm \eta &=\hat{\bm \chi }{{\bm \chi }^{-1}}\\
  & =\left[ \begin{matrix}
   \hat{\bm R}{{\bm R}^{\intercal}} & \hat{\bm v}-\hat{\bm R}{{\bm R}^{\intercal}}\bm{v} & \hat{\bm p}-\hat{\bm R}{{\bm R}^{\intercal}}\bm{p}\\
   \bm{0}_{1\times 3} & 1 & 0  \\
   \bm{0}_{1\times 3} & 0 & 1  \\
\end{matrix} \right]
 \approx \left[ \begin{matrix}
   \bm{I}+\bm \xi _{R}^{\wedge } & {{\bm \xi }_{v}} & {{\bm \xi }_{p}}\\
\bm{0}_{1\times 3} & 1 & 0  \\
   \bm{0}_{1\times 3} & 0 & 1
\end{matrix} \right],
\end{aligned}}
\end{equation}
We define $\bm{\xi}_{IMU}:=\left[ \bm{\xi}^{\intercal}_R,\ \bm{\xi}^{\intercal}_v,\ \bm{\xi}^{\intercal}_p \right]^{\intercal}\in \mathbb{R}^9$. Using $\bm{\xi}_{IMU}$ we can formulate the standard InEKF. However, augmenting the error of IMU biases and camera misalignment will result in an ``imperfect'' InEKF \cite{barrau2015non}, i.e, the state matrix of linearized system (\ref{ode}) becomes state dependent. 

The estimation error of $\bm{R}_c$ is defined by: 
\begin{equation}
\label{caemra_error}
    \eta_c=\hat{\bm{R}}_c\bm{R}_c^{\intercal}=\operatorname{exp}(\bm \xi_{R_c})\approx \bm{I}+\bm{\xi}^{\land}_{R_c}.
\end{equation}
For IMU biases and camera position, the estimation error are calculated by standard vector difference, i.e:
\begin{equation}
\label{bias_error}
    \bm{e}_{b\omega} = \hat{\bm{b}}_{\bm{\omega}}-\bm{b}_{\bm{\omega}},\ \ \ \bm{e}_{ba}= \hat{\bm{b}}_{\bm{a}}-\bm{b}_{\bm{a}},\ \ \ \bm{e}_{pc}=\hat{\bm{p}}_c-\bm{p}_c.
\end{equation}
The linearized estimation error of the entire system could be expressed as the concatenation of the error in (\ref{right_invariant_error}-\ref{bias_error}):
\begin{equation}
\label{full_state}
    \bm{\xi}:=\left[\bm{\xi}^{\intercal}_{IMU} , \bm{e}_{b\omega}^{\intercal}, \bm{e}_{ba}^{\intercal}, \bm{\xi}_{Rc}^{\intercal}, \bm{e}_{pc}^{\intercal}\right]^{\intercal}\in\mathbb{R}^{21}.
\end{equation}
By system dynamics (\ref{bias}-\ref{camera_pose}) and equation (\ref{compute_gut}-\ref{linearize_g}), we could derive the linearized estimation error dynamics:
\begin{equation}
\label{lin_error_dyn}
\small{
\begin{aligned}
& \frac{d}{dt}\left( \bm{\hat{R}}{{\bm{R}}^{\intercal }} \right)\approx {{\left( \bm{\hat{R}}\left( {{\bm{w}}_{\omega }}-{{\bm{e}}_{b\omega }} \right) \right)}^{\times }}, \\
& \frac{d}{dt}\left( \bm{\hat{v}}-\bm{\hat{R}}{{\bm{R}}^{\intercal }}\bm{v} \right)\approx {{\bm{g}}^{\times }}{{\xi }_{R}}+\bm{\hat{R}}\left( {{\bm{w}}_{a}}-{{\bm{e}}_{ba}} \right) \\ 
& +{{{\bm{\hat{v}}}}^{\times }}{{\xi }_{R}}\left( {{\bm{w}}_{\omega }}-{{\bm{e}}_{b\omega }} \right), \\ 
& \frac{d}{dt}\left( \bm{\hat{p}}-\bm{\hat{R}}{{\bm{R}}^{\intercal }}\bm{p} \right)\approx {{\xi }_{v}}+{{{\bm{\hat{p}}}}^{\times }}{\hat{\bm{R}}}\left( {{\bm{w}}_{\omega }}-{{\bm{e}}_{b\omega }} \right) \\ 
& \frac{d}{dt}{{\bm{e}}_{b\omega }}={{\bm{w}}_{b\omega }},\ \ \ \ \frac{d}{dt}{{\bm{e}}_{b\omega }}={{\bm{w}}_{ba}}, \\ 
& \frac{d}{dt}\left( {{{\bm{\hat{R}}}}_{c}}\bm{R}_{c}^{\intercal } \right)=\bm{w}_{Rc}^{\times },\ \ \ \ \frac{d}{dt}{{\bm{e}}_{pc}}={{\bm{w}}_{pc}}. \\
\end{aligned}}
\end{equation}
Representing (\ref{lin_error_dyn}) in matrix form,  we have: 
\begin{equation}
    \dot{\bm{\xi}}=\bm{A}\bm{\xi}+\bm{B}_{\hat{\bm{x}}}\bm{w},
\end{equation}
where,
\begin{equation}
\small{
\label{matrix_A}
    \bm{A}=\left[\begin{array}{cccccc}
\bm{0}_3 & \bm{0}_3 & \bm{0}_3 & -\hat{\bm{R}} & \bm{0}_3 & \bm{0}_{3 \times 6} \\
\bm{g}^{\times} & \bm{0}_3 & \bm{0}_3 & -\hat{\bm{v}}^{\times} \hat{\bm{R}} & -\hat{\bm{R}} & \bm{0}_{3 \times 6} \\
\bm{0}_3 & \bm{I}_{3} & \bm{0}_3 & -\hat{\bm{p}}^{\times} \hat{\bm{R}} & \bm{0}_3 & \bm{0}_{3 \times 6} \\
{\ }&\bm{0}_{12\times 9}&{\ }&{\ }&\bm{I}_{12}&{}
\end{array}\right],}
\end{equation}
\begin{equation}
\small{
\bm{B}_{\hat{\bm{x}}}=\left[\begin{array}{cc} 
\mathrm{Ad}_{\hat{\bm{\chi}}} & \bm{0}_{9 \times 12} \\
\bm{0}_{12 \times 9} & \bm{I}_{12}
\end{array}\right],\mathrm{Ad}_{\hat{\bm{\chi}}}=\left[\begin{array}{ccc}
\hat{\bm{R}} & \bm{0}_3 & \bm{0}_3 \\
-\hat{\bm{v}}^{\times} \hat{\bm{R}} & \hat{\bm{R}} & \bm{0}_3  \\
-\hat{\bm{p}}^{\times} \hat{\bm{R}} & \bm{0}_3 & \hat{\bm{R}}  \\
\end{array}\right],}
\end{equation}
\begin{equation}
    \bm{w}:=\operatorname{vec}\left(\bm{w_{\omega}},\ \bm{w_{a}},\ \bm{0}_{3\times 1},\ \bm{w_{b_{a}}},\ \bm{w_{b_{\omega}}},\ \bm{w_{R_c}},\ \bm{w_{p_c}}\right).
\end{equation}
Note $\bm{w}$ corresponds to the noise and the system covariance is $\bm{P}$. The dynamics of $\bm{P}$ is governed by the Ricatti equation given as follows:
\begin{equation}
    \frac{\rm d}{\rm d t}\bm{P}=\bm{AP}+\bm{PA}^{\intercal}+\bm{Q},\ \ \bm{Q}=\bm{B}_{\hat{\bm{x}}} \operatorname{Cov}(\bm{w}) \bm{B}_{\hat{\bm{x}}}^{\intercal}.
\end{equation}
\subsection {Propagation Steps}
To apply the filter in discrete time, we assume zero-order hold to the input and perform Euler integration from time $t_k$ to $t_{k+1}$. The discrete dynamics becomes:
\begin{equation}
    \hat{\bm{R}}^{-}_{k+1}=\hat{\bm{R}}^{+}_{k} \exp \left(\left(\tilde{\bm{\omega}}_{k}-\hat{\bm{b}}^{+}_{\omega, k}\right) \Delta t\right),
\end{equation}
\begin{equation}
    \hat{\bm{v}}^{-}_{k+1}=\hat{\bm{v}}^{+}_{k}+\hat{\bm{R}}^{+}_{k}\left(\tilde{\bm{a}}_{k}-\hat{\bm{b}}^{+}_{a, k}\right) \Delta t+\bm{g} \Delta t,
\end{equation}
\begin{equation}
    \hat{\bm{p}}^{-}_{k+1}=\hat{\bm{p}}^{+}_{k}+\hat{\bm{v}}^{+}_{k} \Delta t+\frac{1}{2} \hat{\bm{R}}^{+}_{k}\left(\tilde{\bm{a}}_{k}-\hat{\bm{b}}^{+}_{a, k}\right) \Delta t^{2}+\frac{1}{2}\bm{g}\Delta t^{2},
\end{equation}
\begin{equation}
    \hat{\bm{d}}^{-}_{k+1}=\hat{\bm{d}}^{+}_{k},\ \  \hat{\bm{b}}^{-}_{\omega, k+1}=\hat{\bm{b}}^{+}_{\omega, k},\ \ \hat{\bm{b}}^{-}_{a, k+1}=\hat{\bm{b}}^{+}_{a, k},
\end{equation}
\begin{equation}
\hat{\bm{R}}^{-}_{c, k+1}=\hat{\bm{R}}^{+}_{c, k},\ \ \ \ \hat{\bm{p}}^{-}_{c, k+1}=\hat{\bm{p}}^{+}_{c, k}.
\end{equation}
where $\Delta t=t_{k+1}-t_k$,$(\cdot)^{+}_k$ denotes the estimated states at time $t_k$ with all measurements until $t_k$ are processed and $(\cdot)^{-}_k$ denotes the state estimated at time $t_k$ through propagation. 
Given the system matrix (\ref{matrix_A}) at time $t_k$, we have discrete state transformation matrix and discrete-time covariance propagation equation:
\begin{equation}
    \Phi_k=\operatorname{exp_m}\left( \bm{A}_k\Delta t \right),\ \ \ \bm{P}_{k+1}=\Phi_k\bm{P}_k\Phi_k^{\intercal}+{\bm{Q}}_k,
\end{equation}
where ${\bm{Q}}_{k} \approx \Phi_k{\bm{Q}}\Phi_k^{\intercal}\Delta t$. We recommend \cite{hartley2020contact} for more details on the system discretization.
\subsection {Update Steps}
The leg kinematics measurement (\ref{fk_model}) formulates the right-invariant observation \cite{barrau2016invariant}, which exploits the geometry property of the system:
\begin{equation}
    \bm{y}=\bm{\chi}^{-1}\bm{b}+\bm{s},
\end{equation}
where:
\begin{equation}
     \bm{y} = [-\left(\bm{J}_i\dot{\tilde{\bm{\alpha}}}+\tilde{\bm{\omega}}^{\times}\bm{r}_i(\tilde{\bm{\alpha}})\right)^{\intercal},-1,0]^{\intercal},
\end{equation}
\begin{equation}
    \bm{b}=\left[\bm{0}_{1\times 3},-1,0\right]^{\intercal},\ \ \bm s=\left[\bm{n}_f^{\intercal},0,0\right]^{\intercal}.
\end{equation}
As the right-invariant observation only contains elements in $\bm \chi$, we first derive the innovation term for $\bm \chi$ and then expand it to the full state. The update equation for $\bm \chi$ are defined by matrix multiplication:
\begin{equation}
    \hat{\bm \chi}^{+}=\exp\left(\bm{L}\left(\hat{\bm{\chi}}^{-}\bm y-\bm b\right)\right)\hat{\bm{\chi}}^{-},
\end{equation}
\begin{equation}
    \bm{\eta}^{+} = \exp\left(\bm{L}\left(\bm{\eta}^{-}\bm b-\bm b+ \bm{\chi}^{-}\bm s\right) \right)\hat{\bm \eta}^{-},
\end{equation}
where $\bm{L}$ is the observation gain matrix. 
Using the first order approximation of exponential map, we have:
\begin{equation}
    \begin{aligned}
  {{\bm{\eta }}^{+}}&=\exp ({\bm{\xi }_{IMU}^{+}})\approx (\bm{I}+{\bm{\xi}^{+}_{IMU})^{\wedge}} \\ 
 & \approx (\bm{I}+{\bm{\xi}^{-}_{IMU})^{\wedge}}+{{\left( \bm{L\Pi }\left( {{\bm{\eta }}^{-}}\bm{b}-\bm{b}+{{\bm{\chi }}^{-}}\bm{s} \right) \right)}^{\wedge }},  
\end{aligned}
\end{equation}
where $\bm{\Pi}=\left[\bm I_3,\bm 0_{3 \times 2}\right]$ is an auxiliary matrix to select the first three rows of the right hand side. Therefore we have the update equation for $\bm{\xi}_{IMU}$: 
\begin{equation}
\label{update_chi}
\small{
\begin{aligned}
  {{\bm{\xi }}_{IMU}^{+}}={\bm{\xi }_{IMU}^{-}}- \bm{L}\left( \left[\bm{0}_{3},-\bm{I}_3,\bm{0}_{3}\right]{{\bm{\xi}^{-}_{IMU}}}-\left[\bm{0}_{1\times 3}, (\bm{\hat{R}n}_f)^{\intercal},\bm{0}_{1\times 3}\right]^{\intercal} \right).
\end{aligned}
}
\end{equation}
Expand (\ref{update_chi}) to full state (\ref{full_state}), we have:
\begin{equation}
    {{\bm{\xi }}^{+}}={\bm{\xi }^{-}}-\bm{K}\left( \bm{H}{{\bm{\xi}^{-}}}-\left[\bm{0}_{1\times 3}, (\bm{\hat{R}n}_f)^{\intercal},\bm{0}_{1\times 15}\right]^{\intercal} \right),
\end{equation}
where the Kalman gain $\bm{K}$ is given by:
\begin{equation}
    \bm{K}=\bm{PH}^{\intercal}\bm{S}^{-1},\ \ \text{with}\ \bm{S}=\bm{HP^{-}H}^{\intercal}+\bm{N},
\end{equation}
\begin{equation}
\label{constant_h}
\bm{N}=\hat{\bm{R}}\operatorname{Cov}(\bm{n}_f)\hat{\bm{R}}^{\intercal},\ \ \bm{H}=\left[\bm{0}_{3},-\bm{I}_3,\bm{0}_{3\times 15}\right].
\end{equation}
Using the correction term:
\begin{equation}
    \bm{\delta}:=\left[\bm{\delta}^{\intercal}_{IMU} , \bm{\delta}_{b\omega}^{\intercal}, \bm{\delta}_{ba}^{\intercal}, \bm{\delta}_{Rc}^{\intercal}, \bm{\delta}_{pc}^{\intercal}\right]^{\intercal}=\bm{K\Pi}\left(\hat{\bm{\chi}}\bm{y}-\bm{b}\right).
\end{equation}
we finally have the update equations:
\begin{equation}
    \hat{\bm{\chi}}^{+}=\operatorname{exp}\left( \bm{\delta}_{{IMU}}\right)\hat{\bm{\chi}}^{-},
\end{equation}
\begin{equation}
    \bm{b}_a^{+}=\bm{b}_a^{-}+\bm{\delta}_{{ba}},\ \ \ \bm{b}_\omega^{+}=\bm{b}_\omega^{-}+\bm{\delta}_{{b\omega}},
\end{equation}
\begin{equation}
    \bm{R}_c^{+}=\operatorname{exp}\left( \bm{\delta}_{Rc}\right)\bm{R}^{-}_c,\ \ \bm{p}_c^{+}=\bm{p}^{-}_c+\bm{\delta}_{pc},
\end{equation}
\begin{equation}
    \bm{P}^{+}=\left( \bm{I} -\bm{KH}\right)\bm{P}^{-}\left( \bm{I} -\bm{KH}\right)^{\intercal}+\bm{KNK}^{\intercal}.
\end{equation}

The tracking camera measurements (\ref{vel_model}) can not be formulated as a right invariant observation. Therefore, the correction term $\bm{\delta}$ will be computed by standard vector addition. A pseudo measurement based InEKF implementation using a similar update equation has been reported in \cite{AIIMU}. For measurement $\bm{y}=\bm{h}(\bm{x},\bm{n})$, we implement first order approximation to obtain the linearized observation model:
\begin{equation}
\label{obs_model}
\begin{aligned}
\bm{h}(\hat{\bm{x}},\bm{0})-\bm{h}(\bm{x},\bm{n})=\bm{H}_c \bm{\xi}+\bm{Gn}+ h.o.t(\bm{\xi},\bm{n}),
\end{aligned}
\end{equation}
where $\bm{H}_c$ and $\bm{G}$ are Jacobians with respect to the state error and noise term respectively. 

We apply \eqref{obs_model} to \eqref{vel_model} to compute the the observation matrix for the tracking camera measurements:
\begin{equation}
\bm{H}_{c}=\left[\bm{0},\hat{\bm{R}}_c^{\intercal}\hat{\bm{R}}^{\intercal},\bm{0}_{3\times 9}, \bm{\omega}_c^{\times}\hat{\bm{R}}_c^{\intercal}\hat{\bm p}_c^{\times}+\hat{\bm{R}}_c^{\intercal}(\hat{\bm{R}}^{\intercal}\hat{\bm v})^{\times}, -\bm{\omega}_c^{\times}\hat{\bm{R}}^{\intercal}\right].
\label{vary_h}
\end{equation}
Note that we need to take the derivative w.r.t to error defined in \eqref{full_state} to obtain \eqref{vary_h}. The corresponding covariance matrix for this measurements takes the form:
\begin{equation}
    \bm{N}_{c}=\hat{\bm{R}}\hat{\bm{R}}_c\operatorname{Cov}(\bm{\tilde{\bm{v}}}_c)\hat{\bm{R}}_c^{\intercal}\hat{\bm{R}}^{\intercal},
\end{equation}
\begin{equation}
     \operatorname{Cov}(\bm{\tilde{\bm{v}}}_c)=\operatorname{Cov}(\bm{n}_{vc})+(-\hat{\bm{R}}^{\intercal}_c\hat{\bm{p}}_c)^{\times}\operatorname{Cov}(\bm{n}_{\omega c}) (\hat{\bm{R}}^{\intercal}_c\hat{\bm{p}}_c)^{\times}.
\end{equation}
Replacing $\bm{N}$ and $\bm{H}$ in equations (50-51) with $\bm{N}_c$ and $\bm{H}_c$, we have the update term for the tracking camera measurements:
\begin{equation}
    \bm{\delta}_c=\bm{K}\left(\bm{h}(\hat{\bm{x}}) - \tilde{\bm{v}}_c\right).
\end{equation}
It is worth noticing that the observation matrix $\bm{H}$ (\ref{constant_h}) for leg kinematics measurement is a constant while $\bm{H}_c$ (\ref{vary_h}) is state dependent. The state-dependent nature of $\bm{H}_c$ may lead to inconsistency in observability and thus needs more attention.
\subsection{Parameter Tuning}
As an accurate estimation of sensor noise is key to the implementation of filtering, the covariance of the tracking camera measurements should be carefully computed. We define the tracking camera measurement $\bm{\tilde{m}}:=[\bm{\tilde{\omega}}^{\intercal}_c,\bm{\tilde{v}}_c^{\intercal}]^{\intercal}$ and its covariance $\bm{\tilde{w}}_m:=[\bm{n}_{\omega c}^{\intercal},\bm{n}_{vc}^{\intercal}]^{\intercal}$. Since the periodic ground impact while walking causes highly time-varying camera noise, we estimate covariance of $\bm{\tilde{w}}_m$ online rather than using a fixed value. The covariance of $\bm{\tilde{w}}_m$ at time $t_k$ is denoted by $Cov{{({{\bm{w}}_{m}})}_{k}}$, which is estimated by the empirical covariance from $t_{k-n}$ to $t_{k}, n \in \mathbb{N}^{+}$: 
\begin{equation}
\label{emp_cov}
\small{
  \begin{aligned}
  Cov{{({{\bm{w}}_{m}})}_{k}}\approx\frac{1}{n}\sum\limits_{i=0}^{n}{ \bm{e}_{{k-i}} \bm{e}_{{k-i}}^{\intercal}}, \text{with}\ \bm{e}_{{k-i}} = {{{\bm{\tilde{m }}}}_{{{t}_{k-i}}}}-\frac{1}{n}\sum\limits_{j=0}^{n}{{{{\bm{\tilde{m }}}}_{{{t}_{k-j}}}}}.
\end{aligned}
}
\end{equation}
To avoid large delays, we tend to use a small $n$. In the experiment, the filter works well using online tuned $Cov{{({{\bm{w}}_{m}})}_{k}}$ but diverges when we use a fixed value. The estimated covariance is presented in Fig. $\ref{camera_cov}$, which is highly time-varying. 

\section{Observability Analysis}
This section discusses the observability of the filter. As the leg kinematic measurements forms a right-invariant observation, we could obtain the unobservable states of the filter without nonlinear observability analysis \cite{barrau2016invariant}. Similar systems has been thoroughly studied in \cite{ETHslipUKF} and \cite{hartley2020contact}, thus we only analyze the filter with tracking camera measurements. 
\subsection {Nonlinear Observability Analysis}
We employ the notion of locally weak observability depicted in \cite{1977nonlinearobservability}. Consider a nonlinear system in the form:
\begin{equation}
    \begin{aligned}
         \dot{\bm x}&=\bm f(\bm x, \bm u),\ \ \ \ \bm y=\bm h(\bm x).
    \end{aligned}
\end{equation}
Given state ${\bm x}$ and input $\bm u$, the observability matrix is spanned by the gradients of the Lie derivatives:
\begin{equation}
\label{nonlinear_obs}
    \bm{O}({\bm x}, \bm{u})=\left[\begin{array}{c}
\nabla \mathcal{L}_{f}^{0} \bm{h}({\bm x}) \\
\nabla \mathcal{L}_{\bm{f}}^{1} \bm{h}({\bm x}) \\
\vdots
\end{array}\right],
\end{equation}
\begin{equation}
     \mathcal{L}_{f}^{0} \bm{h}({\bm x}) = \bm{h}({\bm x}),...\ \mathcal{L}_{f}^{n} \bm{h}({\bm x}) = \frac{\partial \mathcal{L}_{f}({\bm x})^{n-1} \bm{h}}{\partial \bm{x}}\bm{f},\ \ n\geq1.
\end{equation}
We use exponential coordinates to parameterize rotation:
\begin{equation}
    {\bm{R}}=\exp (\bm{\phi }),\ \ \ \dot{\bm{\phi}}=\bm{R}(\bm\phi)\bm\omega,\ \ \ \bm{\phi}:=\left[\phi_x,\ \phi_y,\ \phi_z \right]^{\intercal},
\end{equation}
where $\bm{\omega}$ could be interpreted as the rotational rate expressed in the robot-frame while $\dot{\bm{\phi}}$ is in the world frame. Similarly, with $\bm{R}_c = \exp(\bm{\phi}_c)$, we then represent the system (\ref{bias}-\ref{camera_pose}) in robot-centric frame to lower the computation burden:
\begin{equation}
\label{robocentric}
\begin{aligned}
\bar{\bm{x}}&:=\left[\begin{array}{c}
\bar{\bm{\phi}}^{\intercal},
\bar{\bm{v}}^{\intercal},
\bar{\bm{p}}^{\intercal},
\bar{\bm{b}}_{\bm{a}}^{\intercal},
\bar{\bm{b}}_{{\omega}}^{\intercal},
\bar{\bm{\phi}}_{c}^{\intercal},
\bar{\bm{p}}_{c}^{\intercal}
\end{array}\right]^{\intercal}\\
&=\left[\begin{array}{c}
\bm{\phi}^{\intercal},
(\bm{R}^{\intercal} \bm{v})^{\intercal},
(\bm{R}^{\intercal} \bm{p})^{\intercal},
\bm{b}_{a}^{\intercal},
\bm{b}_{{\omega}}^{\intercal},
\bm{\phi}_{c}^{\intercal},
\bm{p}_{c}^{\intercal}
\end{array}\right]^{\intercal}.
\end{aligned}
\end{equation}
As the transformation (\ref{robocentric}) is invertible, the observability will not change. The system becomes:
\begin{equation}
\label{system_transformed}
\begin{aligned}
     \dot{\bar{\bm{x}}}=\bm f(\bar{\bm{x}},\bar{\bm u}):=\left[\begin{array}{c}
\bar{\bm{R}}\bar{\bm{\omega}} \\
\bar{\bm{\omega}}^{\times} \bar{\bm{v}}+\bar{\bm{a}}+\bar{\bm{R}}^{\intercal} \bm{g} \\
\bar{\bm{\omega}}^{\times} \bar{\bm{p}}+\bar{\bm{v}} \\
\bm{0}_{12\times 1}
\end{array}\right],\\
\end{aligned}
\end{equation}
\begin{equation}
\label{obs_transformded}
    \bm h_c(\bar{\bm{x}})=\bar{\bm R}^{\intercal}_c\bar{\bm{v}}+\bar{\bm \omega}_{c}^{\times}\bar{\bm R}_c^{\intercal}\bar{\bm p}_c,
\end{equation}
\begin{equation}
    \bar{\bm{\omega}}:=\bm\omega -\bar{\bm b}_{\bm \omega},\ \ \ \bar{\bm{a}}:=\bm a -\bar{\bm b}_{\bm a},\ \ \ \bar{\bm u}:=vec(\bar{\bm{\omega}}, \bar{\bm{a}}).
\end{equation}

Applying (\ref{nonlinear_obs}) to (\ref{system_transformed}, \ref{obs_transformded}) and do row transformation, the observability matrix could be obtained and simplified as:
\begin{equation}
\label{continuous_c}
\small{
    \begin{matrix}
  {\bm{O}_{c}}=\left[ \begin{matrix}
   \bm{0} & \bm{I} & \bm{0} & \bm{0} & \bm{0} & \bm\Delta & {{\left( {{{\bm{\bar{R}}}}_{c}}{{\bm{\omega }}_{c}} \right)}^{\times }}  \\
   {{{\bm{\bar{R}}}}^{\intercal }}{{\bm{g}}^{\times }} & {{{\bm{\bar{\omega }}}}^{\times }} & \bm{0} & {{{\bm{\bar{v}}}}^{\times }} & -\bm{I} & {{\#}_{0,6}} & \bm{0}  \\
   \bm{0} & {{{\bm{\bar{\omega }}}}^{\times 2}} & \bm{0} & {{\#}_{2,4}} & -{{{\bm{\bar{\omega }}}}^{\times }} & {{\#}_{1,6}} & \bm{0}  \\
   {{\#}_{2,1}} & {{{\bm{\bar{\omega }}}}^{\times 3}} & \bm{0} & {{\#}_{3,4}} & -{{{\bm{\bar{\omega }}}}^{\times 2}} & {{\#}_{2,6}} & \bm{0}  \\
   \bm{0} & {{{\bm{\bar{\omega }}}}^{\times 4}} & \bm{0} & {{\#}_{4,4}} & -{{{\bm{\bar{\omega }}}}^{\times 3}} & {{\#}_{3,6}} & \bm{0}  \\
   {{\#}_{4,1}} & {{{\bm{\bar{\omega }}}}^{\times 5}} & \bm{0} & {{\#}_{5,4}} & -{{{\bm{\bar{\omega }}}}^{\times 4}} & {{\#}_{4,6}} & \bm{0}  \\
   \bm{0} & {{{\bm{\bar{\omega }}}}^{\times 6}} & \bm{0} & {{\#}_{6,4}} & -{{{\bm{\bar{\omega }}}}^{\times 5}} & {{\#}_{5,6}} & \bm{0}  \\
\end{matrix} \right] \\ 
\end{matrix},
}
\end{equation}
\begin{equation}
\small{
\begin{aligned}
  & {{\#}_{i,1}}={{{\bm{\bar{\omega }}}}^{\times i}}{{{\bm{\bar{R}}}}^{\intercal }}{{\bm{g}}^{\times }},\ \ \ \ {{{\#}_{i,4}}=\partial {{{\bm{\bar{\omega }}}}^{\times i}}\bm{\bar{v}}}/{\partial {{{\bm{\bar{b}}}}_{\bm{\omega }}}}, \\ 
 & \bm{\Delta}={{{\bm{\bar{v}}}}^{\times }}+{{\left( {{{\bm{\bar{R}}}}_{c}}{{\bm{\omega }}_{c}} \right)}^{\times }}{{\bm{p}}_{c}},\ \ \ {{\#}_{0,6}}={{\left( {{{\bm{\bar{\omega }}}}^{\times }}\bm{\bar{v}}+\bm{\bar{a}}+{{{\bm{\bar{R}}}}^{\intercal }}{{\bm{g}}^{\times }} \right)}^{\times }}, \\ 
 & {{\#}_{i,6}}={{\left( {{{\bm{\bar{\omega }}}}^{\times i}}\left( {{{\bm{\bar{\omega }}}}^{\times }}\bm{\bar{v}}+\bm{\bar{a}}+\tfrac{1+{{(-1)}^{i}}}{2}{{{\bm{\bar{R}}}}^{\intercal }}{{\bm{g}}^{\times }} \right) \right)}^{\times }}. \\ 
\end{aligned}
}
\end{equation}
Without the effect of noise, we also have $\bar{\bm{\omega}}^{\times}=({{{\bm{\bar{R}}}}_{c}}{{\bm{\omega }}_{c}})^{\times }$. Therefore we could determine the null space of the observability matrix, i.e. $\bar{\bm{U}}_c\ s.t.\ \bar{\bm{O}}_c\bar{\bm{U}}_c=\bm 0$: 
\begin{equation}
\small{
    {\bm{U}}_c=\left[ \begin{matrix}
   {{\bm{g}}^{\intercal }}  & 0 & 0 & 0 & 0 & 0 & 0  \\
   \bm{0} & \bm{0} & \bm{I} & \bm{0} & \bm{0} & \bm{0} & \bm{0}  \\
   0 & 0 & 0 & 0 & 0 & 0 & {{{\bm{\bar{\omega }}}}^{\intercal }}  \\
\end{matrix} \right]^{\intercal }.
}
\end{equation}
The first two rows of ${\bm{U}}_c^{\intercal}$ suggest that the orientation around the gravity vector and the position are unobservable. The position of the camera along the body angular velocity is also unobservable with only one measurement according to the last row. Note, the camera position will be fully observable when multiple different measurements are obtained.

The observability matrix may lose rank with certain robot motion, see the simulation of camera pose estimation in the video. We next list the singular cases based on the angular and linear velocity of robot pelvis. If $\bm{\bar{\omega }}=0, \bm{\bar{v}}=0$, the last 2 columns of ${\bm{O}}_c$ in \eqref{continuous_c} become 0 thus rank loss is 5. In this case, the camera pose becomes fully unobservable. If $\bm{\bar{\omega }}=0, \bm{\bar{v}}\neq 0$, the last column of ${\bm{O}}_c$ becomes zero, while the rank of the last but one column depends on $\bar{\bm{v}}^{\times}$. Thus the total rank loss is 3. In this case, the orientation around the velocity vector and camera position are unobservable.
\subsection {Discrete Observability Analysis}
The filter is applied in linearized and discretized version:
\begin{equation}
\label{time-varying-system}
    \begin{aligned}
 {{\bm{x}}_{k+1}}={{\bm{\Phi }}_{k}}{{\bm{x}}_{k}},\ \ \ \ {{\bm{y}}_{k}}={{\bm{H}}_{k}}{{\bm{x}}_{k}}.
\end{aligned}
\end{equation}
However, observability of this discrete linear time-varying system may differ from the underlying nonlinear system as the linearized points deviate from the real states. This inconsistency in EKF-SLAM has been analyzed and solved by Observability Constraint EKF \cite{OC-EKF}. Similar problems in EKF design for legged robot navigation are also studied in \cite{ETHSlipPrior}.
Following the method in \cite{OC-EKF}, we analyze the local observability matrix \cite{time-varying-matrix} at the ``standard'' operating point, i.e, linearized at the latest estimated value:
\begin{equation}
\label{discrete-matrix}
\small{
    \hat{\bm{O}}=\left[ \begin{matrix}
   \bm{H}_{k}^{-}  \\
   \bm{H}_{k+1}^{-}\bm{\Phi }_{k}^{+}  \\
   \bm{H}_{k+2}^{-}\bm{\Phi }_{k+1}^{+}\bm{\Phi }_{k}^{+}  \\
   \vdots   \\
\end{matrix} \right].
}
\end{equation}
By applying (\ref{discrete-matrix}), the discrete observability matrix could be obtained and simplified to:
\begin{equation}
\label{discrete_c}
\small{
    \begin{matrix}
  {{\bm{\hat{O}}}_{c}}\text{=}\left[ \begin{matrix}
   \bm{0} & \bm{R}_{k}^{-\intercal } & \bm{0} & \bm{0} & \bm{0} & {{\#}_{0,6}} & \bm{R}_{c,k}^{-}\bm{\omega }_{c,k}^{\times }\bm{R}_{c,k}^{-\intercal }  \\
   \bm{R}_{k+1}^{-\intercal }{{\bm{g}}^{\times }}\Delta t & {{\#}_{1,2}} & \bm{0} & {{\#}_{1,4}} & {{\#}_{1,5}} & {{\#}_{1,6}} & {{\#}_{1,7}}  \\
   {{\#}_{2,1}} & {{\#}_{2,2}} & \bm{0} & {{\#}_{2,4}} & {{\#}_{2,5}} & {{\#}_{2,6}} & {{\#}_{2,7}}  \\
   {{\#}_{3,1}} & {{\#}_{3,2}} & \bm{0} & {{\#}_{3,4}} & {{\#}_{3,5}} & {{\#}_{3,6}} & {{\#}_{3,7}}  \\
   {{\#}_{4,1}} & {{\#}_{4,2}} & \bm{0} & {{\#}_{4,4}} & {{\#}_{4,5}} & {{\#}_{4,6}} & {{\#}_{4,7}}  \\
   {{\#}_{5,1}} & {{\#}_{5,2}} & \bm{0} & {{\#}_{5,4}} & {{\#}_{5,5}} & {{\#}_{5,6}} & {{\#}_{5,7}}  \\
   {{\#}_{6,1}} & {{\#}_{6,2}} & \bm{0} & {{\#}_{6,4}} & {{\#}_{6,5}} & {{\#}_{6,6}} & {{\#}_{6,7}}  \\
\end{matrix} \right] \\ 
\end{matrix}
}
\end{equation}
\begin{equation}
\small{
    \begin{aligned}
  & {{\#}_{i,1}}=i\bm{R}_{k+i}^{-\intercal }{{\bm{g}}^{\times }}\Delta t,i\ge 1,{{\#}_{i,2}}=\bm{R}_{k+i}^{-\intercal },i\ge 1 \\ 
 & {{\#}_{1,4}}=-\bm{R}_{k+1}^{-\intercal }\left( \bm{v}_{k}^{+\times }\bm{R}_{k}^{+}\Delta t+\tfrac{1}{2}{{\bm{g}}^{\times }}\bm{R}_{k}^{+}\Delta {{t}^{2}} \right) \\ 
 & {{\#}_{i,4}}=-\bm{R}_{k+i}^{-\intercal }\sum\nolimits_{j=0}^{i-1}{\left( \bm{v}_{k+j}^{+\times }\bm{R}_{k+j}^{+}\Delta t+\tfrac{1}{2}{{\bm{g}}^{\times }}\bm{R}_{k+j}^{+}\Delta {{t}^{2}} \right)}- \\ 
 & \bm{R}_{k+i}^{-\intercal }\sum\nolimits_{l=0}^{i-2}{(i-l-1){{\bm{g}}^{\times }}\bm{R}_{k+l}^{+}\Delta {{t}^{2}}},i\ge 2 \\ 
 & {{\#}_{i,5}}=-\sum\nolimits_{j=0}^{i-1}{\bm{R}_{k+i}^{-\intercal }\bm{R}_{k+j}^{+}\Delta t},i\ge 2,{{\#}_{i,7}}=\bm{R}_{c,i}^{-}\bm{\omega }_{c,i}^{\times }\bm{R}_{c,i}^{-\intercal },i\ge 1 \\ 
 & {{\#}_{i,6}}=\bm{R}_{k+i}^{-\intercal }\bm{v}_{k+i}^{\times }+\bm{\omega }_{c,k+i}^{\times }\bm{R}_{c,k+i}^{-\intercal }\bm{p}_{c,k+i}^{-\times },i\ge 0 \\ 
\end{aligned}
}
\end{equation}
As the third column of $\bm{g}^{\times}$ is always 0, the first column of the $\hat{\bm O}_c$ suggests that $\phi_z$ is always non-observable despite the discrepancy between the operating point and real state. {\color{black}The position is also unobservable according to the third column.}

{\color{black}When ${\bm{\omega}_c}$ is time-varying, the camera position is fully observable as multiple different measurements are considered. In this case, the last column of $\hat{\bm O}_c$ has full rank, but essentially not conflicting with ${\bm O}_c$. When ${\bm{\omega}_c}$ is strictly a non-zero constant for at least 7 consecutive time steps, the rank of the last column of $\hat{\bm O}_c$ is greater than ${\bm O}_c$. As the unobservable state depends on the real input that is inaccessible, it is impossible to design the operating points like \cite{ETHSlipPrior} or \cite{OC-EKF}. However, this case is very rare as the robot motion is highly dynamical. Therefore, our filter will stay consistent with the underlying nonlinear system except for this rare case.} 
\section{Experimental Validation}
We launched experiments on the Cassie bipedal robot to evaluate the proposed filter. The Cassie robot is equipped with an IMU and 14 encoders. The IMU can provide angular velocity and linear acceleration in robot-frame at 800 Hz. The encoders can provide joint angles and their derivatives are numerically computed at 2000Hz. Contact states are obtained via spring deflection measured by encoders and forward kinematics. A Intel Realsense T265 Tracking camera is mounted on the top of robot to provide velocity and angular velocity at 200Hz. Motion capture system is setup to obtain the ground truth.

\begin{figure}
  \centering
  \includegraphics[width=1\columnwidth]{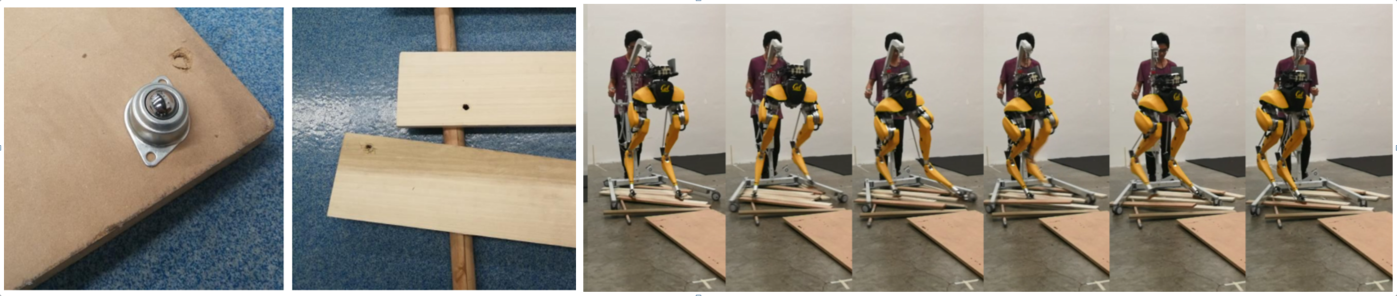}
  \caption{Cassie robot walking on slippery and unstable terrain.}
  \label{exp_shots}
   \vspace{-13pt}
\end{figure}

\begin{figure}
  \centering
  \includegraphics[width=1\columnwidth]{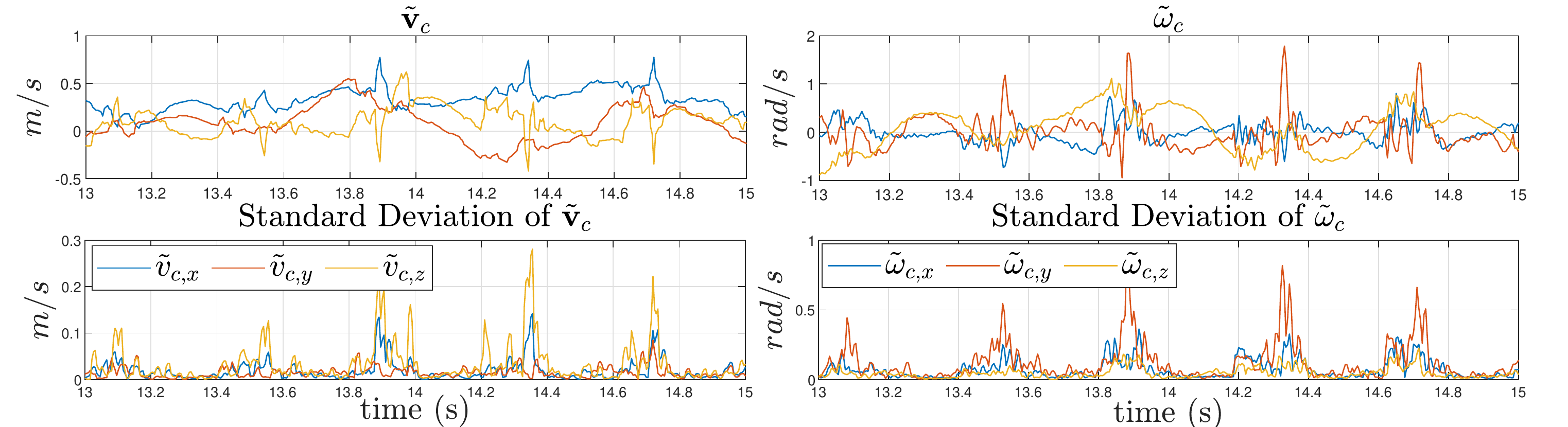}
  \caption{Tracking camera measurements \& empirical std. dev. We use $n=5$ in equation \eqref{emp_cov}.}
  \label{camera_cov}
\end{figure}

The robot is walking over slippery terrain for about 4 meters within 35 seconds with the controller from \cite{controller} implemented. Snapshots of experiments are presented in Fig. \ref{exp_shots}. Tracking camera measurements and their empirical standard deviations are presented in Fig. \ref{camera_cov}. We also plotted the states estimated by fusion of only inertial and leg kinematics, denoted by legend ``camera off'' for comparison. We applied the outlier detection method \cite{ETHslipUKF} in both cases. The kinematics measurement with Mahalanobis distance greater than a threshold $\rho$, i.e $\chi^2={{\bm{\delta}}^{\intercal }}{{\bm{S}}^{-1}}\bm{\delta}>\rho$ will be discarded. As the slippage persisted for multiple seconds, the estimated value will diverge if we discard all the slippage when the camera is off. Thus, we carefully tune the threshold to maintain convergence while trying to discard as much outliers as possible. We use $\rho=30.1$.

\begin{figure}
   \centering
   \includegraphics[width=1\columnwidth]{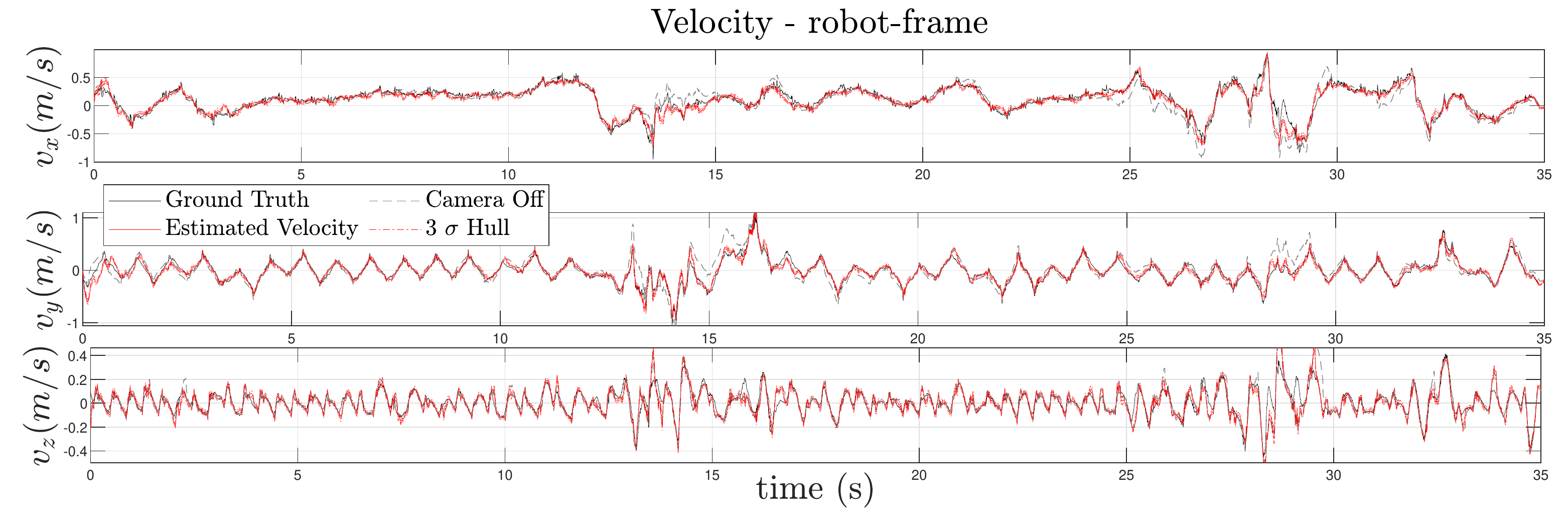}
   \caption{Velocity estimation in robot-frame. The RMSE for $v_x$, $v_y$ and $v_z$ are 0.0703 m/s, 0.0660 m/s and 0.0513 m/s. }
   \label{vb}
    \vspace{-10pt}
 \end{figure}
\begin{figure}
   \centering
   \includegraphics[width=1\columnwidth]{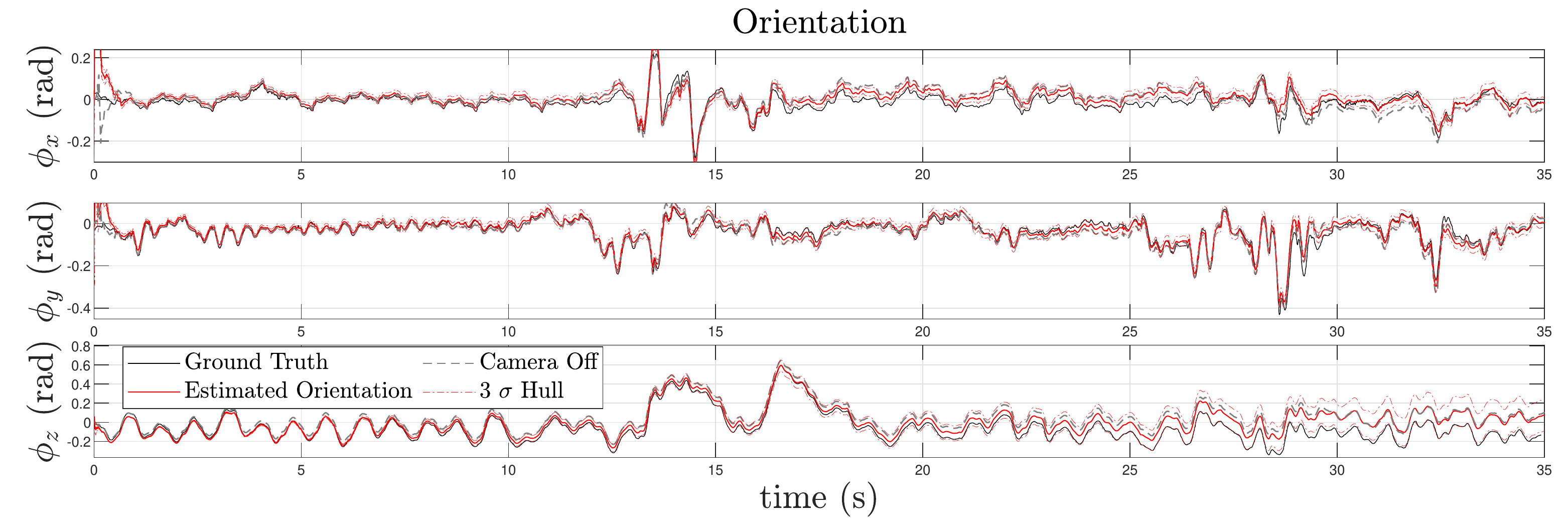}
     \caption{Estimated robot orientation. The RMSE of $\phi_x$, $\phi_y$ and $\phi_z$ are 0.0362 rad, 0.0213 rad and 0.185 rad respectively.}
   \label{theta}
 \end{figure}

The estimated robot velocity and orientation in the robot-frame are presented in Fig. \ref{vb} and \ref{theta} respectively. The velocity in the robot-frame is observable thus converging fast. Rotation around the gravity axis, i.e, $\phi_z$ is not observable, thus the uncertainty slowly grows. $\phi_x$ and $\phi_y$ are observable and converges after a transient state. Due to outliers, the inertial-kinematics method inaccurately estimated some points but our method is always consistent with the ground truth.

\begin{figure}
  \centering
  \includegraphics[width=1\columnwidth]{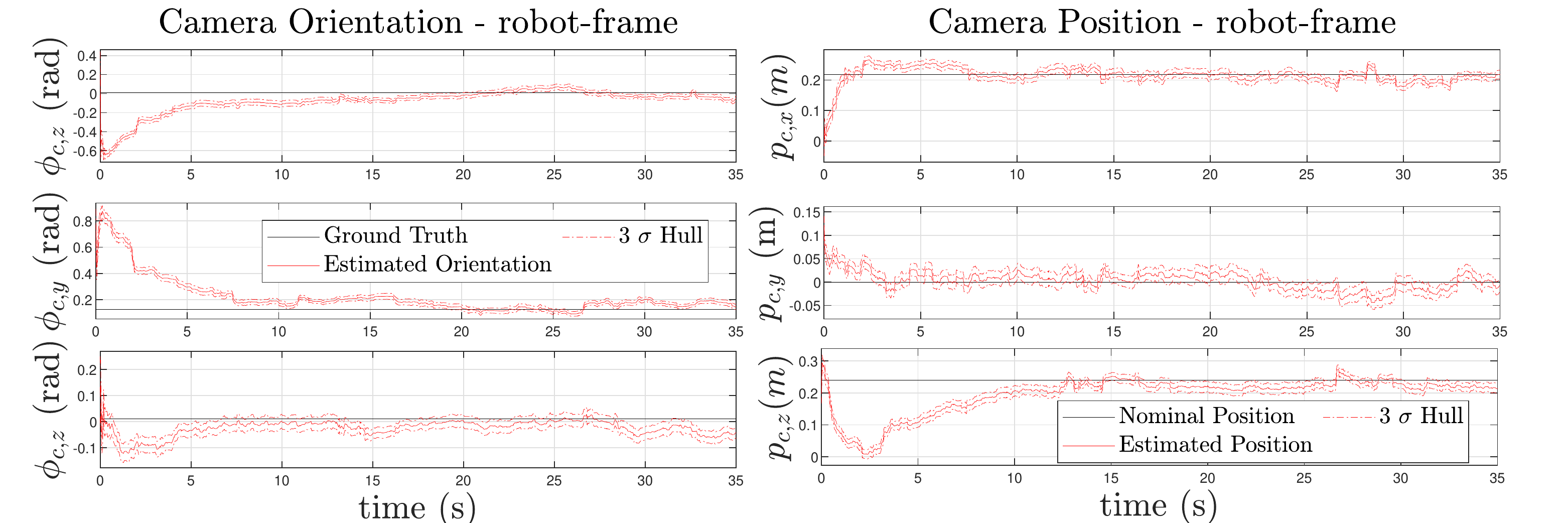}
  \caption{Estimated camera pose. The values converge to the calibrated value after a transient state. The ground truth camera orientation is calibrated by aligning the robot and camera IMU. The nominal camera position is obtained by the CAD model of the platform.}
  \label{cam_rot}
\end{figure}

\begin{figure}
  \centering
  \includegraphics[width=1\columnwidth]{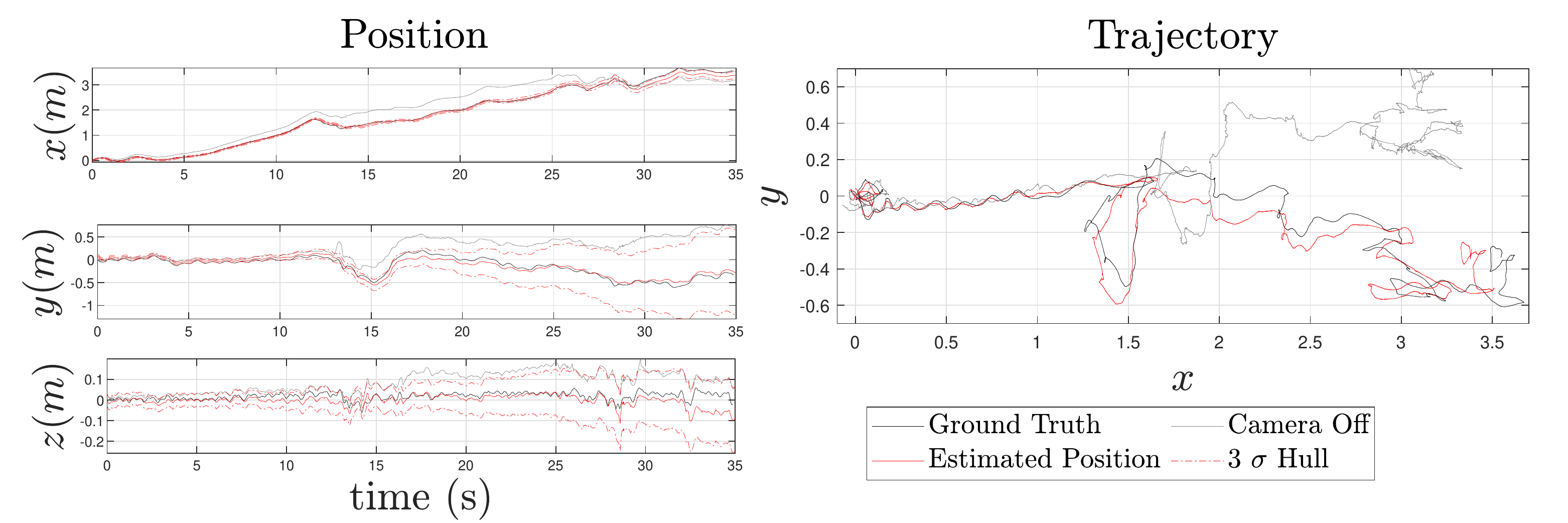}
  \caption{Estimated robot position in 3D space. The estimated trajectory overlapped well with ground truth before the major slip at $x \approx 1.5m$. A shift is seen after the slip but the trajectory did not diverge. Note that the robot position is unobservable. }
  \label{Position_long}
\end{figure}

Fig. \ref{Position_long} presents the estimated Cartesian position. As the position of robot is unobservable, a discrepancy is seen after an initial drift. However, the overall horizontal drift is smaller than $5\%$ over the 3.5 m trajectory. The vertical drift is as small as 0.07 m, which is satisfactory compared to the InEKF in \cite{hartley2020contact}. The orientation and position of camera in the robot-frame are presented in Fig. \ref{cam_rot}. All the 6 quantities converge and the $3 \sigma$ covariance hull overlapped well with the ground truth. This result is quite satisfactory considering the highly time-varying camera measurement noise.



\section{Conclusions}
This paper proposed a state estimation approach for legged robots by fusing inertial, velocity measurements from a tracking camera as well as from leg kinematics. The obtained information is fused by an invariant extended Kalman filter. The leg kinematics based velocity estimation is formulated as right-invariant observation. The misalignment between the camera and the robot-frame is also modeled thus enables auto-calibration of the camera pose. An online covariance tuning method is proposed to handle the highly time-varying measurement noise from the tracking camera. Nonlinear and discrete observability analysis suggest that only the rotation around the gravity vector and the absolute robot position is not observable, and our proposed filter remains consistent with the underlying nonlinear system.

Our method is evaluated in legged locomotion experiments with significant amount of slippage involved. The result suggests that our method could accurately estimate the robot inclination and the robot velocity with robustness. Although absolute velocity and rotation about gravity is unobservable, their drifts remains small. Camera pose is also successfully estimated by the filter. 




\section*{Acknowledgement}
We would like to thank B. Zhang, P. Kotaru and S. Chen for experiment and M. Brossard for discussions on InEKF theory. We would also like to extend special thanks to the High Performance Robotics Lab as well as Michigan's Bipedal Robotics Lab.



\balance

\bibliographystyle{ieeetrans.bst}
\bibliography{ref}
\end{document}